\documentclass[11pt,a4paper]{article}
\usepackage[hyperref]{emnlp2018}
\usepackage{times}
\usepackage{latexsym}

\usepackage{url}
\usepackage{CJK}

\usepackage{graphicx}
\usepackage{subfigure}
\usepackage{booktabs}       
\usepackage{amsfonts}       
\usepackage{nicefrac}       
\usepackage{comment}
\usepackage{amsmath}
\usepackage{bbm}
\usepackage{booktabs}
\usepackage{multirow}
\aclfinalcopy 

\title{Beyond Error Propagation in Neural Machine Translation: \\
Characteristics of Language Also Matter}

\author{Lijun Wu$^1$\thanks{Authors contribute equally to this work.}, Xu Tan$^2$\footnotemark[1], Di He$^3$, Fei Tian$^2$, Tao Qin$^2$, Jianhuang Lai$^1$ \and Tie-Yan Liu$^2$\\
$^1$School of Data and Computer Science, Sun Yat-sen University\\
\quad$^2$Microsoft Research \\
\quad$^3$Key Laboratory of Machine Perception, MOE, School of EECS, Peking University \\
wulijun3@mail2.sysu.edu.cn;\;stsljh@mail.sysu.edu.cn;\;\\
\{xuta,fetia,taoqin,tyliu\}@microsoft.com;\; di\_he@pku.edu.cn\;
}

\date{}

\begin{document}
\maketitle
\begin{abstract}
Neural machine translation usually adopts autoregressive models and suffers from exposure bias as well as the consequent error propagation problem. Many previous works have discussed the relationship between error propagation and the \emph{accuracy drop} (i.e., the left part of the translated sentence is often better than its right part in left-to-right decoding models) problem. In this paper, we conduct a series of analyses to deeply understand this problem and get several interesting findings. (1) The role of error propagation on accuracy drop is overstated in the literature, although it indeed contributes to the accuracy drop problem. (2) Characteristics of a language play a more important role in causing the accuracy drop: the left part of the translation result in a right-branching language (e.g., English) is more likely to be more accurate than its right part, while the right part is more accurate for a left-branching language (e.g., Japanese). Our discoveries are confirmed on different model structures including Transformer and RNN, and in other sequence generation tasks such as text summarization.

\end{abstract}

\section{Introduction}
Neural machine translation (NMT) has attracted much research attention in recent years~\cite{bahdanau2014neural,DBLP:conf/naacl/ShenTHQL18,DBLP:conf/coling/SongTHLQL18,DBLP:conf/icml/XiaTTQYL18,he2016dual,wu2017adversarial,rl_study}. The major approach to the task typically leverages an encoder-decoder framework \cite{DBLP:conf/emnlp/ChoMGBBSB14,DBLP:conf/nips/SutskeverVL14} and the decoder usually generates the target tokens one by one from left to right autoregressively, in which the generation of a target token is conditioned on previously generated target tokens.

It has been observed that for an NMT model with left-to-right decoding, the right part words in its translation results are usually worse than the left part words in terms of accuracy~\cite{DBLP:journals/corr/abs-1801-05122,DBLP:conf/nips/BengioVJS15,DBLP:journals/corr/RanzatoCAZ15,DBLP:journals/corr/abs-1803-05567,DBLP:conf/naacl/LiuUFS16,DBLP:conf/aaai/LiuFUS16}. This phenomenon is referred to as \textit{accuracy drop} in this paper. A straightforward explanation to accuracy drop is  \textit{error propagation}: If a word is mistakenly predicted during inference, the error will be propagated and the future words conditioned on this one will be impacted. Different methods have been proposed to address the problem of accuracy drop~\citep{DBLP:conf/aaai/LiuFUS16,DBLP:conf/naacl/LiuUFS16,DBLP:journals/corr/abs-1803-05567}.

Instead of solving the problem, in this paper, we aim to deeply understand the causes of the problem. In particular, we want to answer the following two questions:


\begin{itemize}
\item Is error propagation the main cause of accuracy drop?
\item Are there any other causes leading to accuracy drop?
\end{itemize}

To answer these two questions, we conduct a series of experiments to analyze the problem.

First, we train NMT models separately using left-to-right and right-to-left decoding~\cite{DBLP:conf/wmt/SennrichHB16,DBLP:conf/naacl/LiuUFS16,he2017decoding,gao2018efficient}  on several language pairs (i.e., German to English, English to German, and English to Chinese). If error propagation is the main cause of accuracy drop, then the right part words in the translation results generated by right-to-left NMT models should be more accurate than the left part words. However, we observe the opposite phenomenon that the accuracy of the right part words of the translated sentences in both left-to-right and right-to-left models is lower than that of the left part, which contradicts with error propagation. This shows that error propagation alone cannot well explain the accuracy drop and even suggests that error propagation may not exist or matter.

Second, to further investigate the influence of error propagation on accuracy drop, we conduct a set of experiments with teacher forcing~\cite{DBLP:journals/neco/WilliamsZ89} during inference, in which we feed the ground-truth preceding words to predict the next target word. Teacher forcing eliminates exposure bias as well as error propagation in inference. The results verify the existence of error propagation, since the later part (the right part in left-to-right decoding and the left part in right-to-left decoding) of the translation results get more accuracy improvement with teacher forcing, regardless of the decoding direction. Meanwhile, the accuracy of the right part is still lower than that of the left part with teacher forcing, which demonstrates that there must be some other causes apart from error propagation leading to accuracy drop.

Third, inspired by linguistics, we find that the concept of branching ~\cite{berg2011structure,payne2006exploring} can help to explain the problem. We conduct the third set of experiments to study the correlation between language branching and accuracy drop.
We find that if a target language is right branching such as English, the accuracy of the left part words is usually higher than that of the right part words, no matter for left-to-right or right-to-left NMT models, while for a left-branching target language such as Japanese, the accuracy of the left part words is usually lower than that of the right part, no matter for which models.
The intuitive explanation is that a right-branching language has a clearer structure pattern (easier to predict) in the left part of sentence than that in the right part, since the main subject of the sentence is usually put in the left part. We calculate two statistics to verify this assumption: n-gram statistics (including n-gram frequency and conditional probabilities) and dependency parsing statistics. For right-branching languages, we found higher n-gram frequency/conditional probabilities as well as more dependencies in the left part compared with that in the right part. The opposite results are also found in left-branching languages.

We summarize our findings as follows.

\begin{itemize}
\item Through empirical analyses, we find that the influence of error propagation is overstated in the literature, which may misguide the future research. Error propagation alone cannot fully explain the accuracy drop in the left or right part of sentence.
\item We find the branching in linguistics well correlates with accuracy drop in the left or right part of sentence and the corresponding analysis on n-gram and dependency parsing statistics well explain this phenomenon.
\end{itemize}


Our studies show that linguistics can be very helpful to understand existing machine learning models and build better models for language related tasks. We hope that our work can bring some insights to the research on neural machine translation. We believe that our findings can help us to design better translation models. For example, the finding on language branching suggests us to use left-to-right NMT models for right-branching languages such as English and right-to-right NMT models for left-branching languages such as Japanese.

\section{Related Work}
\label{related_work}

\subsection{Exposure Bias and Error Propagation}
Exposure bias and error propagation are two different concepts but often mentioned together in literature~\cite{DBLP:conf/nips/BengioVJS15,DBLP:conf/acl/ShenCHHWSL16,DBLP:journals/corr/RanzatoCAZ15,DBLP:conf/naacl/LiuUFS16,DBLP:conf/aaai/LiuFUS16,DBLP:journals/corr/abs-1801-05122,DBLP:journals/corr/abs-1803-05567}. Exposure bias refers to the fact that the sequence generation model is usually trained with teacher-forcing while generates the sequence autoaggressviely during inference. This discrepancy between training and inference can yield errors that accumulate quickly along the generated sequence, which is known as error propagation~\cite{DBLP:conf/nips/BengioVJS15,DBLP:conf/acl/ShenCHHWSL16,DBLP:journals/corr/RanzatoCAZ15}.

~\citet{DBLP:conf/nips/BengioVJS15} propose the scheduled sampling method to eliminate the exposure bias and the resulting error propagation, which achieves promising performance on sequence generation tasks such as image captioning. ~\citet{DBLP:conf/acl/ShenCHHWSL16,DBLP:journals/corr/RanzatoCAZ15} improve the basic maximum likelihood estimation (MLE) with reinforcement learning or minimum risk training and aim to address the limitation of MLE training and exposure bias problem.

\subsection{Tackling Accuracy Drop}

~\cite{DBLP:conf/naacl/LiuUFS16,DBLP:conf/aaai/LiuFUS16,DBLP:journals/corr/abs-1801-05122,DBLP:journals/corr/abs-1803-05567} mainly ascribe accuracy drop (the accuracy of right part words is worse than that in the left part in most cases) to error propagation and propose different methods to solve this problem. ~\citet{DBLP:conf/naacl/LiuUFS16,DBLP:conf/aaai/LiuFUS16,DBLP:journals/corr/abs-1803-05567} use agreement regularization between the left-to-right and right-to-left models to achieve better performance. ~\citet{DBLP:journals/corr/abs-1801-05122} and~\cite{DBLP:journals/corr/abs-1803-05567} propose to use two-pass decoding to refine the generated sequence to yield better quality.

All these works focus on error propagation and accuracy drop. To our knowledge, there is no deep study about other causes of accuracy drop. In this paper, we aim to conduct such a study. Our study shows that accuracy drop is not only caused by error propagation, but also the characteristics of language itself.



\section{Error Propagation and Accuracy Drop}

\subsection{Error Propagation is Not the Only Cause}
\label{sec_3_measure}
A left-to-right NMT model feeds target tokens one by one from left to right in training and generate target tokens one by one from left to right during inference, while a right-to-left NMT model trains and generates token in the reverse direction. Intuitively, if error propagation is the root cause of accuracy drop, then a right-to-left NMT model will generate translations with better right half accuracy than the left half. In this section, we study the results of both left-to-right and right-to-left NMT models to analyze the relationship between error propagation and accuracy drop.

We conduct experiments on three translation tasks with different language pairs, which include: IWSLT 2014 German-English (De-En), WMT 2014 English-German (En-De) and WMT 2017 English-Chinese (En-Zh). We choose the state-of-the-art NMT model Transformer~\cite{DBLP:conf/nips/VaswaniSPUJGKP17} as the basic model structure and train two separate models with left-to-right and right-to-left decoding on each language pair. More details about the datasets and model descriptions can be found in supplementary materials (section A.1 and A.2). We evenly split each generated sentence into the left half and the right half with same number of words\footnote{1) For most of the sentences, the last word of the sentence is period which is easy to decode. To make a fair comparison, we simply remove the last period before dividing the translation sentence. 2) For sentence with an odd number of words, we simply remove the word in the middle position to make the left half and right half have the same number of words.}. Then for both the left and right half, we compute their accuracy with respect to the reference target sentence, in terms of BLEU score~\cite{DBLP:conf/acl/PapineniRWZ02} \footnote{We use the \emph{multi-bleu.perl} script \url{https://github.com/moses-smt/mosesdecoder/scripts/generic/multi-bleu.perl}. When computing BLEU score of the left or right half, the reference is the full reference sentence.}.

\begin{table}[!tbp]
\small
\centering 
\begin{tabular}{ l  c  c  c  c  c} 
\toprule
& De-En & & En-De & & En-Zh   \\
\midrule
left-to-right & 31.42 & & 26.93 & & 20.79  \\
right-to-left & 30.00 & & 25.35 & & 20.23  \\
\bottomrule
\end{tabular}
\caption{BLEU scores on the test set of the three translation tasks with both left-to-right and right-to-left decoding.}
\label{no_feed_full}
\end{table}

\begin{table}[!tbp]
\small
\centering 

\begin{tabular}{ l  c  c  c  c  c } 
\toprule
\textbf{left-to-right}  & De-En & & En-De & & En-Zh \\
\midrule
Left & \textbf{10.17} & & \textbf{7.90} & & \textbf{7.41} \\
Right & 8.39 & & 6.60 & & 5.91 \\

\bottomrule\bottomrule
\textbf{right-to-left} & De-En & & En-De & & En-Zh  \\
\midrule
Right & 7.83 &  & 6.45 &  & 5.77   \\
Left & \textbf{9.41} &  & \textbf{7.11} &  & \textbf{7.01}    \\
\bottomrule
\end{tabular}
\caption{BLEU scores of the left and right half of left-to-right and right-to-left NMT models. In \cite{DBLP:conf/aaai/LiuFUS16}, the authors report the \emph{partial BLEU} score without length penalty, our result is consistent with \emph{partial BLEU} if simply removing length penalty when calculating BLEU.}
\label{r2l_no_feed}
\end{table}

We first report the BLEU scores of the full translation results (without split) in Table~\ref{no_feed_full}. As can be seen, the accuracy of the model is comparable to state-of-the-art results~\cite{DBLP:conf/nips/VaswaniSPUJGKP17,DBLP:conf/wmt/WangCJYCLSWY17,Wang2018Dual}. Afterwards we report the BLEU scores of the left half and the right half in Table~\ref{r2l_no_feed}. We have several observations.

\begin{itemize}
\item When translating from left-to-right, the BLEU score of the left half is higher than the right half on all the three tasks, which is consistent with previous observation and is able to be explained via error propagation.

\item When translating from right-to-left, the accuracy of the left half (in this way it's the later part of the generated sentence) is still higher than the right half. Such an observation is contradictory to the previous analyses between error propagation and accuracy drop, which regard that accumulated error brought by exposure bias will deteriorate the quality in later part of translation (i.e., the left half).
\end{itemize}

The inconsistent observation above suggests that error propagation is not the only cause of accuracy drop that there are other factors beyond error propagation for accuracy drop. It even challenges the existence of error propagation: does error propagation really exist? In the next section we try to answer this question through teacher forcing experiments.

\subsection{The Influence of Error Propagation}
\label{teacher_forced}

Teacher forcing~\cite{DBLP:journals/neco/WilliamsZ89} in sequence generation means that when training a sequence generation model, we feed the previous ground-truth tokens as inputs to predict the next target word. Here we apply teacher forcing in the inference phase of NMT: to generate the next word $\hat{y_i}$, we input the preceding ground-truth words $y_{<i}$ rather than the previously generated words $\hat{y}_{<i}$, which largely alleviates the effect of error propagation, since there will be no error propagated from the previously generated words.

\begin{table}[!tbp]
\small
\centering 
\begin{tabular}{ l  c  c  c  c  c  c  c  } 
\toprule
\multirow{3}{*}{\textbf{De-En}}   & \multicolumn{3}{c}{\textbf{left-to-right}} & & \multicolumn{3}{c}{\textbf{right-to-left}} \\
\cmidrule{2-4}  \cmidrule{6-8}
 & 0 & 1 & $\Delta$ & & 0 & 1 & $\Delta$  \\
\midrule
  Left & \textbf{10.17} & \textbf{10.71} & 0.54 & & \textbf{9.41} & \textbf{10.41} &  1.00\\
Right & 8.39 & 9.25 &  0.86 & & 7.83  & 8.45 & 0.62  \\
\bottomrule\bottomrule
\multirow{3}{*}{\textbf{En-De}}  & \multicolumn{3}{c}{\textbf{left-to-right}} & & \multicolumn{3}{c}{\textbf{right-to-left}} \\
\cmidrule{2-4}  \cmidrule{6-8}
  & 0 & 1 & $\Delta$ & & 0 & 1 & $\Delta$  \\
\midrule
Left & \textbf{7.90} & \textbf{9.43} & 1.53 & & \textbf{7.11}  & \textbf{10.71} & 3.60\\
Right & 6.60 & 8.36 &  1.76 & & 6.45  & 8.37  & 1.92   \\
\bottomrule\bottomrule
\multirow{3}{*}{\textbf{En-Zh}}  & \multicolumn{3}{c}{\textbf{left-to-right}} & & \multicolumn{3}{c}{\textbf{right-to-left}} \\
\cmidrule{2-4}  \cmidrule{6-8}
  & 0 & 1 & $\Delta$ & & 0 & 1 & $\Delta$  \\
\midrule
Left & \textbf{7.41} & \textbf{9.11} & 1.70 & & \textbf{7.01} & \textbf{9.83} &2.82   \\
Right & 5.91 & 8.55 & 2.64 & & 5.77  & 7.54 & 1.77  \\

\bottomrule
\end{tabular}
\caption{BLEU scores. "0" represents the translation results without teacher forcing during inference, and "1" represents the translation results with teacher forcing during inference. $\Delta$ represents the BLEU score improvement of teacher forcing over normal translation.}
\label{feed}
\end{table}

Same as last section, we evaluate the quality of the left and right half of the translation results generated by both the left-to-right and right-to-left models. The results are summarized in Table \ref{feed}. For comparison, we also include the BLEU scores of normal translation (without teacher forcing). We have several findings from Table~\ref{feed} as follows:

\begin{itemize}
\item Exposure bias exists. The accuracy of both left and right half tokens in the normal translation is lower than that in  teacher forcing, which feeds the ground-truth tokens as inputs. This demonstrates that feeding the previously generated tokens (which might be incorrect) in inference indeed hurts translation accuracy.

\item Error propagation does exist. We find the error is accumulated along the sequential generation of the sentence. Taking En-Zh and the left-to-right NMT model as an example, the BLEU score improvement of the right half (the second half of the generation) of teacher forcing over normal translation is 2.64, which is much larger than the accuracy improvement of the left half (the first half of the generation): 1.70. Similarly, for En-Zh with the right-to-left NMT model, the BLEU score improvement of the left half (the second half of the generation) of teacher forcing over normal translation is 2.82, which is much larger than the accuracy improvement of the right half (the first half of the generation): 1.77.
\item Other causes exist. Taking En-De translation with the left-to-right model as an example, the accuracy of the left half (9.43) is higher than that of the right half (8.36) when there is no error propagation with teacher forcing. Similar results can be found in other language pairs and models. This suggests that there must be some other causes leading to accuracy drop, which will be studied in the next section.
\end{itemize}

\section{Language Branching Matters}
\label{language}
Section~\ref{sec_3_measure} and~\ref{teacher_forced} together show that error propagation has influence on but is not the only cause of accuracy drop. We hypothesize that the language itself, i.e., its characteristics, may explain the phenomenon of accuracy drop.

\citet{DBLP:conf/coling/WatanabeS02} finds that left-to-right decoding performs better for Japanese-English translation while right-to-left decoding performs better for English-Japanese translation. We conduct the same analysis settings as in Section~\ref{sec_3_measure} and~\ref{teacher_forced} on English-Japanese (En-Jp) translation dataset. More details about this dataset and model descriptions can be found in supplementary materials (section A.1 and A.2).

Table~\ref{no_feed_jp} shows the BLEU score on the En-Jp test set. It can be observed that regardless of decoding direction (i.e., from left-to-right or from right-to-left) and with or without teacher forcing, the accuracy of the right half is always higher than that in the left half. This observation on Japanese is opposite to English, German and Chinese in Section~\ref{sec_3_measure} and~\ref{teacher_forced}, and motivates us to investigate the differences between these languages.

\begin{table}[!tbp]
\small
\centering 
\begin{tabular}{ l  c  c  c  c  c  c  c  c  } 
\toprule
& \multicolumn{2}{c}{\textbf{left-to-right}} & & \multicolumn{2}{c}{\textbf{right-to-left}}  \\
\cmidrule{2-3}  \cmidrule{5-6}
  & 0 & 1 & & 0 & 1  \\
\midrule
left & 7.90 & 9.91 & & 7.45 & 8.95   \\
right & \textbf{8.70} & \textbf{11.52} & & \textbf{9.24} & \textbf{10.59} \\
\bottomrule
\end{tabular}
\caption{BLEU scores on En-Jp test set. "0" represents the normal translation results, and "1" represents the teacher-forcing translation results.}
\label{no_feed_jp}
\end{table}

We find that a linguistics concept, the branching, can differentiate Japanese from other languages such as English/German. Branching refers to the shape of the parse trees that represent the structure of sentences~\cite{berg2011structure,payne2006exploring}. Usually, right-branching sentences are head-initial, which means the main subject of the sentence is described first, and is followed by a sequence of modifiers that provide additional information about the subject. On the contrary, left-branching sentences are head-final that putting such modifiers to the left of the sentence~\cite{payne2006exploring}.

English is a typical right-branching language, while Japanese is almost fully left-branching~\cite{wiki:Head-directionality_parameter}. The two languages demonstrate the opposite phenomenon of accuracy drop  as shown in previous studies. When we say a language is typical left/right-branching, we mean most of the sentences in this language follows the left/right-branching structure. While being predominantly right-branching, German is less conclusively so than English. Chinese features a mixture of head-final and head-initial structures, with the noun phrases are head-final while the strict head/complement ordering sentences are head-initial as right-branching~\cite{wiki:Head-directionality_parameter}, but less conclusively than German.

We believe the language branching is a main cause of accuracy drop. Intuitively, the main subject of a right-branching sentence is described first (in the left part) and is followed by additional modifiers (in the right part) \cite{berg2011structure}. Therefore, the left half of a right-branching sentence is more likely to possess a clearer structure pattern and thus lead to higher generation accuracy than in the right part, since the main subject is usually simpler and clearer than the modifiers that providing additional information about the subject. In next section, we will verify this intuition this assumption from a statistical perspective.



\section{Correlation between Language Branching and Accuracy Drop}
As previous work~\cite{DBLP:conf/icml/ArpitJBKBKMFCBL17} shows, neural networks are easy to learn and memorize simple patterns but difficult to make a correct prediction on noise examples. In this section, we study different branching languages from two aspects, including the n-gram statistics of a target language, which has been used as a kind of characterization of hardness of learning~\cite{bengio2009curriculum}, and the dependency statistics in parse trees. We show that these statistics well correlate with the accuracy drop between the left half and the right half of translation results.

\subsection{N-gram Statistics}
Intuitively speaking, if a pattern occurs frequently and deterministically, it is easy to be learned by neural networks. By comparing the general statistics on the n-gram frequency and n-gram conditional probability of the left and right half tokens, we link the language branching to accuracy drop.

Denote a bilingual dataset $D=\{(x_i,y_i\}$, $i=1,\cdots,M$, where each $y_i$ is a sequence of words $y_i=\{y_i^1,\cdots,y_i^{T_i}\}$, $T_i$ is the length of $y_i$. $F^{l}_{i,n}$ and $P^{l}_{i,n}$ denote the average n-gram frequency and n-gram conditional probability of the left half of $y_i$~\footnote{Again, we assume $T_i$ is an even number. If not, we simply remove the middle word of $y_i$, as done in Section \ref{sec_3_measure}.}, i.e.,
\begin{equation}
\small
\begin{aligned}
\label{eq_n_gram_freq}
F^{l}_{i,n}&=\frac{1}{T_i/2-n+1} \sum_{j=1}^{T_i/2-n+1} F(y_i^{j},...,y_i^{j+n-1}), \\
P^{l}_{i,n}&=\frac{1}{T_i/2-n+1} \sum_{j=1}^{T_i/2-n+1} P(y_i^{j+n-1} | y_i^{j},...,y_i^{j+n-2}),
\end{aligned}
\end{equation}
where $F(.)$ and $P(.)$ are the n-gram frequency and n-gram conditional probability calculated from the training dataset. Similarly, $F^{r}_{i,n}$ and $P^{r}_{i,n}$ denote the n-gram frequency and n-gram conditional probability of the right half.


We calculate the average n-gram frequencies $F^{l}_{n}$ and $F^{r}_{n}$ of the left half and right half over all the target sentences in the training set. We also calculate the average n-gram conditional probabilities $P^{l}_{n}$ and $P^{r}_{n}$ over all the training sentences to compare the uncertainty of phrases in the left half and right half.
\begin{equation}
\small
\begin{aligned}
\label{eq_n_gram_freq}
F^{l}_{n} &=\frac{1}{M}\sum^{M}_{i=1} F^{l}_{i,n}, ~ F^{r}_{n} =\frac{1}{M}\sum^{M}_{i=1} F^{r}_{i,n}, \\
P^{l}_{n} &= \frac{1}{M} \sum^{M}_{i=1} P^{l}_{i,n}, ~ P^{r}_{n} =\frac{1}{M} \sum^{M}_{i=1} P^{r}_{i,n}. \\
\end{aligned}
\end{equation}

We also calculate the ratio of the sentences that the frequency/conditional probability of left half is bigger/smaller than that in the right half, denoted as $RF^{l>r}_{n}$/$RF^{l<r}_{n}$ and $RP^{l>r}_{n}$/$RP^{l<r}_{n}$:

\begin{equation}
\small
\begin{aligned}
\label{eq_n_gram_prob}
RF^{l>r}_{n} &=\frac{1}{M}\sum^{M}_{i=1} \mathbbm{1}\{F^{l}_{i,n} > F^{r}_{i,n}\},  \\
RF^{l<r}_{n} &=\frac{1}{M}\sum^{M}_{i=1} \mathbbm{1}\{F^{l}_{i,n} < F^{r}_{i,n}\}, \\
RP^{l>r}_{n} &=\frac{1}{M}\sum^{M}_{i=1} \mathbbm{1}\{P^{l}_{i,n} > P^{r}_{i,n}\},  \\
RP^{l<r}_{n} &=\frac{1}{M}\sum^{M}_{i=1} \mathbbm{1}\{P^{l}_{i,n} < P^{r}_{i,n}\}. \\
\end{aligned}
\end{equation}

We choose $n=2$ and $3$ to calculate the metrics in Equation~\ref{eq_n_gram_freq} and~\ref{eq_n_gram_prob} on different translation datasets. The numbers are listed in Table~\ref{table_ngram_freq} and~\ref{table_ngram_prob}.

We can see the 2/3-gram frequency as well as the conditional probability of the left half is higher than that of the right half for right-branching languages including English, German and Chinese in De-En, En-De and En-Zh translation datasets. For left-branching language Japanese, the result is opposite. The n-gram frequency and conditional probability statistics are consistent with our observations on accuracy drop in left/right-branching languages and verify our hypothesis: right-branching languages have clearer patterns in left part (with larger n-gram frequency as well as the conditional probability) and consequently leads to higher translation accuracy in the left part than the right part; left-branching languages are opposite.


\begin{table}[t]
\centering
\small
\begin{tabular}{l c c c c c}
\toprule
& \multicolumn{2}{c}{\textbf{De-En}} & & \multicolumn{2}{c}{\textbf{En-De}}   \\
\cmidrule{2-3}  \cmidrule{5-6}

&  \textbf{2-gram} &  \textbf{3-gram} & & \textbf{2-gram} & \textbf{3-gram} \\
\midrule
$F^{l}_{n}$ &  \textbf{5713.8} & \textbf{3122.7} & &  \textbf{13811.8} & \textbf{687.1} \\
$F^{r}_{n}$ &  3026.5 & 1377.6 & & 11692.2 & 419.9 \\
\midrule
$RF^{l>r}_{n}$  & \textbf{59.6\%} & \textbf{55.8\%} & &  \textbf{53.8\%} & \textbf{53.6\%} \\
$RF^{l<r}_{n}$  & 38.8\% & 37.6\% & & 46.0\% & 45.0\% \\
$\Delta$ & 20.8\% & 18.2\% &  &  7.8\% & 8.6\%  \\
\bottomrule
\bottomrule
& \multicolumn{2}{c}{\textbf{En-Zh}} & & \multicolumn{2}{c}{\textbf{En-Jp}} \\
\cmidrule{2-3}  \cmidrule{5-6}
&  \textbf{2-gram} &  \textbf{3-gram} &  & \textbf{2-gram} &  \textbf{3-gram}  \\
\midrule
$F^{l}_{n}$  & \textbf{17707.0} & \textbf{1954.1} &  & 18910.0 & 1350.0 \\
$F^{r}_{n}$  & 16256.4 & 1250.5 &  & \textbf{21076.7} & \textbf{1754.0} \\
\midrule
$RF^{l>r}_{n}$  & \textbf{51.9\%} & \textbf{50.2\%} &  & 41.2\% & 38.0\% \\
$RF^{l<r}_{n}$  & 46.7\% & 43.9\% &  & \textbf{51.7\%} & \textbf{52.3\%} \\
$\Delta$ & 5.2\% & 6.3\% &  &  -10.5\% & -14.3\%  \\
\bottomrule
\end{tabular}
\caption{The n-gram frequency statistics on different translation datasets. $F^{l}_{n}$ and $F^{r}_{n}$ represent the average of n-gram frequency of left and right half of target sentences. $RF^{l>r}_{n}$ and $RF^{l<r}_{n}$ represent the ratio that the n-gram frequency of left half of sentences are bigger/smaller than that of the right half. $\Delta=RF^{l>r}_{n}-RF^{l<r}_{n}$. Note that the sum of $RF^{l>r}_{n}$ and $RF^{l<r}_{n}$ is less than 1 since sentence with less than 4 words does not contribute to the n-gram statistics.}
\label{table_ngram_freq}
\end{table}


\begin{table}[!tpb]
\centering
\small
\begin{tabular}{l c c c c c}
\toprule
& \multicolumn{2}{c}{\textbf{De-En}} & & \multicolumn{2}{c}{\textbf{En-De}}   \\
\cmidrule{2-3}  \cmidrule{5-6}

& \textbf{2-gram} &  \textbf{3-gram} & &\textbf{2-gram} & \textbf{3-gram} \\
\midrule
$P^{l}_{n}$ & \textbf{0.137} & \textbf{0.181} & & \textbf{0.082} & \textbf{0.155} \\
$P^{r}_{n}$ & 0.092 & 0.116 & & 0.080 & 0.148 \\
\midrule
$RP^{l>r}_{n}$ & \textbf{59.8\%} & \textbf{56.6\%} & & \textbf{50.6\%} & \textbf{51.7\%} \\
$RP^{l<r}_{n}$ & 38.7\% & 36.4\% & & 49.2\% & 47.0\% \\
$\Delta$ &  21.2\%  &  20.2\% &  & 1.4\% &  4.7\% \\
\bottomrule
\bottomrule
& \multicolumn{2}{c}{\textbf{En-Zh}} & & \multicolumn{2}{c}{\textbf{En-Jp}} \\
\cmidrule{2-3}  \cmidrule{5-6}

& \textbf{2-gram} &  \textbf{3-gram} & &\textbf{2-gram} & \textbf{3-gram} \\
\midrule
$P^{l}_{n}$ & \textbf{0.064} & \textbf{0.113} & & 0.082 & 0.171 \\
$P^{r}_{n}$ & 0.055 & 0.108 & & \textbf{0.086} & \textbf{0.191} \\
\midrule
$RP^{l>r}_{n}$ & \textbf{52.1\%} & \textbf{47.8\%} & & 43.9\% & 39.4\% \\
$RP^{l<r}_{n}$ & 46.6\% & 47.0\% &  & \textbf{49.2\%} & \textbf{50.9\%} \\
$\Delta$ &  5.5\%  & 0.8\% &  & -5.3\% &  -11.5\% \\
\bottomrule
\end{tabular}
\caption{The n-gram conditional probability statistics on different translation datasets. $P^{l}_{n}$ and $P^{r}_{n}$ represent the average n-gram conditional probability of left and right half of target sentences. $RP^{l>r}_{n}$ and $RP^{l<r}_{n}$ represent the ratio that the n-gram frequency of left half are bigger/smaller than that of the right half. $\Delta=RP^{l>r}_{n}-RP^{l<r}_{n}$. Note that the sum of $RP^{l>r}_{n}$ and $RP^{l<r}_{n}$ is less than 1 due to two reasons: (1) sentence with less than 4 words does not contribute to the statistics, and (2) we remove the n-gram condition probability with the denominator less than 100 to make probability calculation robust.}
\label{table_ngram_prob}
\end{table}


\begin{figure*}
\centering
\subfigure[Accuracy drop v.s 3-gram frequency gap (\%).]{
\begin{minipage}[b]{0.48\textwidth}
\includegraphics[width=1\linewidth]{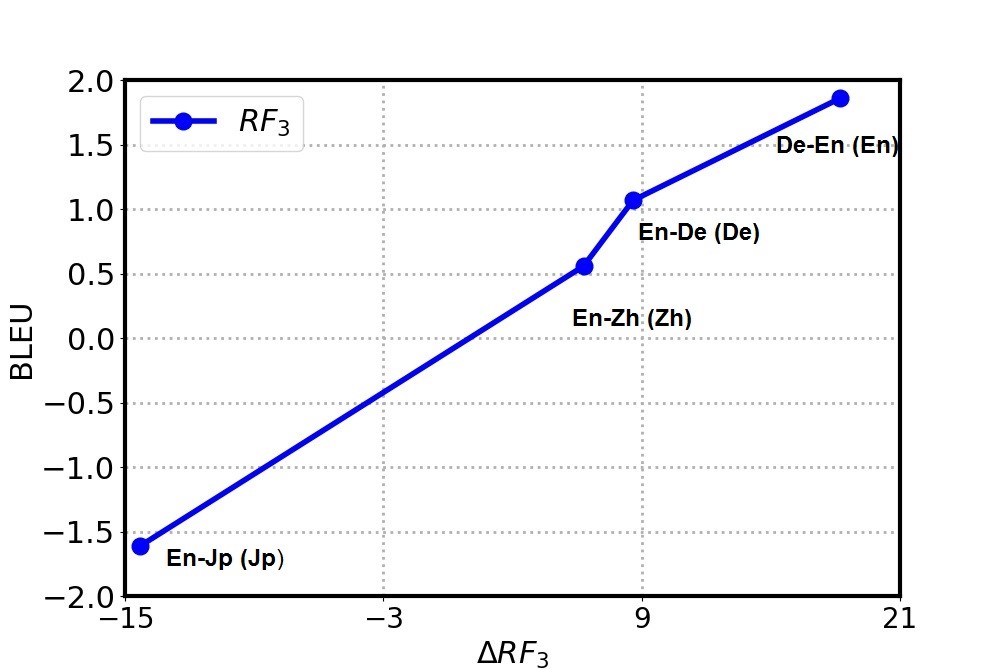}
\label{fig:3-gram_freq}
\end{minipage}
}
\subfigure[Accuracy drop v.s 3-gram conditional probability gap (\%).]{
\begin{minipage}[b]{0.48\textwidth}
\includegraphics[width=1\linewidth]{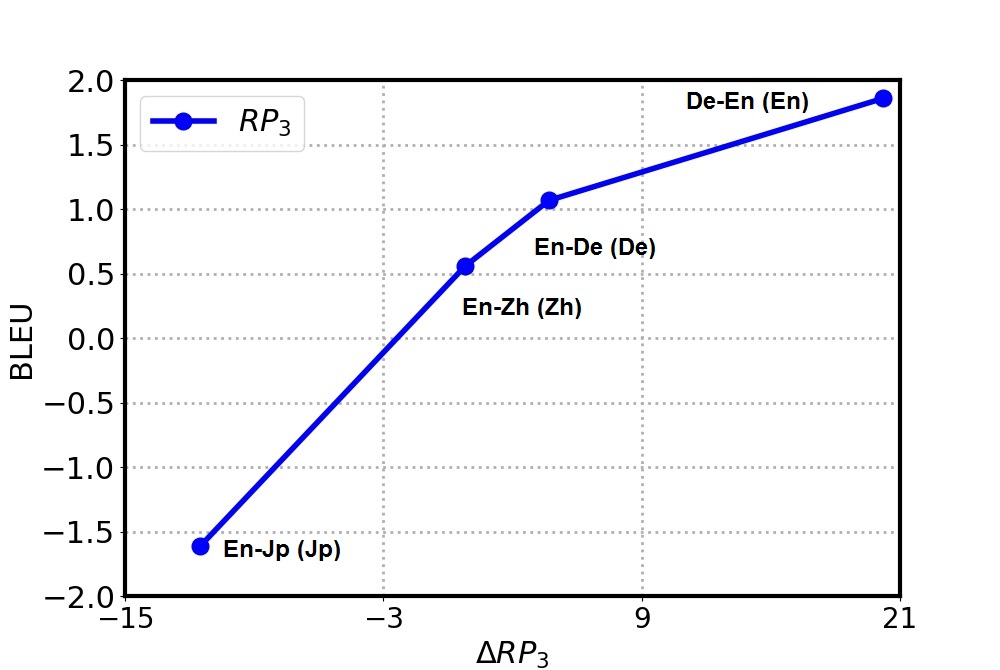}
\label{fig:3-gram_prob}
\end{minipage}
}
\caption{Accuracy drop (the gap between the left/right BLEU score) with respect to the $\Delta RF_3$ and $\Delta RP_3$ from Table~\ref{table_ngram_freq} and~\ref{table_ngram_prob} in the four translation tasks. The x-axis $\Delta RF_3$ and $\Delta RP_3$ represent the gap of between the left and right ratio of the 3-gram frequency/conditional probability defined in Table \ref{table_ngram_freq} and \ref{table_ngram_prob}. The y-axis represents the accuracy drop in terms of BLEU score calculated by the teacher forcing decoding.}
\label{fig:delta_bleu}
\end{figure*}

We further visualize how the accuracy drop (between the left half and right half of the translations) correlates with the gap of n-gram statistics in the left and right part. The accuracy drop (e.g., BLEU score) of left/right half is taken from the teacher-forcing with left-to-right decoding in Table~\ref{feed}, and the n-gram gap is taken from the $\Delta$ in the last row of Table~\ref{table_ngram_freq} and~\ref{table_ngram_prob}. Figure \ref{fig:delta_bleu} shows strong correlation between accuracy drop and the gap of n-gram statistics: As the gap of n-gram statistics increases from negative values to positive values, the accuracy drop also increases from negative to positive.

\subsection{Dependency Statistics}
In this subsection, we study language branching from the perspective of dependency structure.  We hypothesize that if the left/right half of sentence contains more dependencies between its intra words, this half should be easier to predict, leading to higher accuracy. Here we analyze the English sentence in De-En translation and Japanese sentence in En-Jp translation, since English is fully right-branching and Japanese is fully left-branching as introduced before.

For English parsing, we utilize the well-acknowledged Standford Parser\footnote{\url{https://nlp.stanford.edu/software/lex-parser.shtml}} to parse the sentences. After obtaining the parsing results, we split the sentence into left and right half, and separately count the numbers of dependencies in each half\footnote{For simplicity, we just count the number of dependency, without considering dependency types. The detailed parsing formats can be found in the supplementary material (Section A.3).}. For Japanese, we leverage the open-source toolkit J.DepP\footnote{\url{http://www.tkl.iis.u-tokyo.ac.jp/~ynaga/jdepp/}} to parse the sentence, and then count the number of dependencies of each half.

\begin{table}[tbp]
\small
\centering 
\begin{tabular}{ c  c  c  } 
\toprule 
  & \textbf{English} & \textbf{Japanese} \\
\cmidrule{2-3}
Left & \textbf{40242} & 921735 \\
Right  & 31509 & \textbf{1570630} \\
\bottomrule
\end{tabular}
\caption{Number of dependencies in left and right half of English (De-En) and Japanese (En-Jp) training corpus. The number varies a lot since the two training corpus have different training sentences.}
\label{parsing}
\end{table}

We provide the results in Table \ref{parsing}. As can be observed, for English sentences, the left-half words depend more on each other than the right-half words, while for the Japanese sentences, the right-half words have more dependencies. This observation is consistent with our observations on accuracy drop, and can well explain the high accuracy of left part in English translation and right part in Japanese translation.

\section{Extended Analyses and Discussions}
\label{extended_analysis}
We have analyzed the accuracy drop problem from the view of error propagation and language itself in previous sections. In this section, we further provide extended analyses and several discussions to give a more clear understanding of the accuracy drop problem.

\subsection{More Languages on Left-Branching}
The previous analyses are based on four languages, three right-branching (En, De, Zh) and one left-branching language (Jp). To avoid the experimental bias and randomness, we provide one more translation task, English-Turkish (En-Tr) translation\footnote{The detailed dataset and model description can be found in supplementary material (section A.1 and section A.2).}, as Turkish is a  left-branching language. We simply calculate the BLEU score of the left/right half in left-to-right and right-to-left decodings, as in Section \ref{sec_3_measure} and \ref{teacher_forced}.

\begin{table}[tbp]
\small
\centering 
\begin{tabular}{ c  c  c } 
\toprule 
  & 0 & 1 \\
\midrule
Left & \textbf{5.83} & 7.44   \\
Right  & 5.27 & \textbf{7.96}  \\
\bottomrule
\end{tabular}
\caption{BLEU scores on En-Tr test set with left-to-right generation. Normal translation is denoted as ``0'', and teacher-forcing translation is denoted as ``1''.}
\label{En-Tr}
\end{table}

The result is provided in Table~\ref{En-Tr}. For the left-to-right decoding, the accuracy of the left half is higher than that of the right half in the normal translation. However, the accuracy of the right half becomes higher with teacher forcing translation. This demonstrates that English-Turkish translation performs similar to English-Japanese translation as the accuracy of right half is higher than that of the left half. But different from what we observed in Japanese, Turkish shows the opposite phenomenon: the influence of language branching is weaker than error propagation.

\subsection{Other Model Structures}
\label{structure}
One may wonder whether the results in the paper are biased towards a certain model structure as we use Transformer on all the above analyses. To address such concerns, we conduct an additional experiment on De-En translation task with RNN (GRU)-based model\footnote{The detailed setting for GRU based RNN model can be found in supplementary material (section A.2).}. The results are shown in Table \ref{rnn_model} and the observations are consistent with what we observed on Transformer. The accuracy of the left half of the De-En translation sentence is always higher than the right half, in both the left-to-right and right-to-left decodings.

\begin{table}[tbp]
\small
\centering 
\begin{tabular}{ c  c  c  } 
\toprule 
  & \textbf{left-to-right} & \textbf{right-to-left} \\
\midrule
Full & 27.63 & 25.44 \\
\midrule
Left & \textbf{9.17} & \textbf{8.37} \\
Right  & 7.51 & 7.25 \\
\bottomrule
\end{tabular}
\caption{BLEU scores on the left-to-right and right-to-left translation sentences on the De-En test set, with RNN-based model. ``Full" means the BLEU score of the whole translation sentence.}
\label{rnn_model}
\end{table}

\subsection{Other Sequence Generation Tasks}
We conduct experimental analysis on abstractive summarization, which is also a sequence generation task. The goal of the task is to recap a long news sentence into a short summary. We use Gigaword dataset which contains $3.8M$ training pairs, $190k$ validation and $2k$ test pairs of English sentence, and train an RNN-based model for sentence summarization. The accuracy is measured by the commonly used metric ROUGE F1 score and are reported in Table \ref{summarization}.

We observe the same phenomenon as in translation tasks. The accuracy of the left half is always better than the right half, no matter in left-to-right or right-to-left decoding, since the target language English is a right-branching language. 

\begin{table}[tbp]
\centering
\small
\begin{tabular}{l c c c }
\toprule
& \multicolumn{3}{c}{\textbf{left-to-right}}  \\
\cmidrule{2-4}

& \textbf{ROUGE-1} & \textbf{ROUGE-2} &  \textbf{ROUGE-L}\\
\midrule
Full & 35.55 & 16.66 & 33.01 \\
\midrule
Left & \textbf{24.44} & \textbf{9.87} & \textbf{23.34} \\
Right & 21.31 & 8.32 & 20.38  \\
\bottomrule
 & \multicolumn{3}{c}{\textbf{right-to-left}}   \\
 \cmidrule{2-4}

 & \textbf{ROUGE-1} & \textbf{ROUGE-2} & \textbf{ROUGE-L} \\
\midrule
Full & 35.22 & 16.55 & 32.59  \\
\midrule
Right & 21.62 & 8.41 & 20.48 \\
Left & \textbf{23.60} & \textbf{9.54} & \textbf{22.52}  \\

\bottomrule

\end{tabular}
\caption{ROUGE F1 scores for left-to-right and right-to-left generated translation sentences in abstractive summarization task. ROUGE-N stands for N-gram based ROUGE F1 score, ROUGE-L stands for longest common subsequence based ROUGE F1 score. ``Full" means the entire translation sentence.}
\label{summarization}
\end{table}

\section{Conclusion}
\label{conclusion}
In this work, we studied the problem of accuracy drop between the left half and the right half of the results generated by neural machine translation models. We found the influence of error propagation is overstated in literature and error propagation alone cannot explain accuracy drop. We showed that language branching well correlates to the accuracy drop problem and the evidences on n-gram statistics as well as the dependency statistics well support this correlation. Our discoveries suggest that left-to-right NMT models fit better for right-branching languages (e.g., English) and right-to-left NMT models fit better for left-branching languages (e.g., Japanese).

For future works, we will study more left/right-branching languages as well as other languages that have no obvious branching characteristics. We will also investigate how language branching influences other natural language tasks, especially for neural networks based models.

\bibliography{emnlp2018}

\newpage
\appendix

\section{Beyond Error Propagation in Neural Machine Translation: Characteristics of Language Also Matter (Supplemental Material)}

\subsection{NMT Datasets}
The translation datasets we used in our experiments are from five different translation tasks. The details are in the following descriptions.

1) IWSLT 2014 German-English (De-En)~\cite{Cettolo2014Report} translation task. The dataset contains about $153k$ parallel training sentences, and $6.7k$ sentences for both validation and test set. 2) WMT 2014 English-German (En-De) translation task. The dataset contains about $4.5M$ training pairs\footnote{https://nlp.stanford.edu/projects/nmt/}, $6k$ validation set and $3k$ test set. 3) WMT 2017 English-Chinese (En-Zh) translation task\footnote{http://www.statmt.org/wmt17/translation-task.html}. There are nearly $24M$ sentences in the training set, $2k$ for both validation and test. 4) ASPEC English-Japanese (En-Jp) \cite{nakazawa2016aspec} translation, this corpus contains $1.5M$ training samples, nearly $1.8k$ for validation and test set. 5) IWSLT 2014 English-Turkish (En-Tr)~\cite{Cettolo2014Report} translation dataset, which contains about $350k$ training pairs, $16k$ valid pairs and $7.4k$ test pairs.

For the first three translations, the sentences are preprocessed using byte-pair encoding \cite{DBLP:conf/acl/SennrichHB16a} into sub-words, while for the En-Jp translation, the sentences are on the word level. For the EN-Tr translation, the dataset is separately processed into morphological segmentation by using Zemberek\footnote{\url{https://github.com/orhanf/zemberekMorphTR}}.

\subsection{NMT Models}
\paragraph{Transformer Model} The generation model we used is Transformer~\cite{DBLP:conf/nips/VaswaniSPUJGKP17}, which is based on the self-attention architecture. We use \emph{transofmer\_small} setting for De-En and En-Tr, \emph{transformer\_base\_v1} for En-De and En-Jp, \emph{transformer\_big} for En-Zh~\cite{DBLP:journals/corr/abs-1803-07416}. For the right-to-left model, we simply reverse the target language sentence as our training data. For example, for De-En translation, we first reverse the target English sentence, and then align the original source German sentence together with reversed English sentence as pair data for training. The models are optimized through Adam as used in the original paper \cite{DBLP:conf/nips/VaswaniSPUJGKP17}. During decoding phase, we generate the translation sentence by simply greedy search.

\paragraph{RNN Model} We also conduct experiments on RNN based models. The RNN models we adopted in Section 5 are GRU based single-layer models, which contain a bidirectional GRU encoder and a unidirectional GRU decoder. For the De-En translation task, the GRU model is a relatively small model, for which the embedding size and hidden size are both set as $256$. For the summarization task, the embedding size of the GRU model is $512$ and the hidden size is $1024$. The models are trained by Adadelta with learning rate $1.0$.

\subsection{Dependency Parsing Results}
The dependency parsing results for English corpus and Japanese corpus are provided here, the examples are as follows.

\paragraph{English parsing} For English parsing we use Stanford Parser\footnote{\url{https://nlp.stanford.edu/software/lex-parser.shtml}} together with NLTK \cite{NLTK}:

CASE: \emph{``and the great indicator of that , of course , is language loss"}

PARSING:

\emph{\textbf{[}((u'loss', u'NN'), u'cc', (u'and', u'CC')), \\
((u'loss', u'NN'), u'nsubj', (u'indicator', u'NN')), \\
((u'indicator', u'NN'), u'det', (u'the', u'DT')), \\
((u'indicator', u'NN'), u'amod', (u'great', u'JJ')), \\
((u'indicator', u'NN'), u'prep', (u'of', u'IN')), \\
((u'of', u'IN'), u'pobj', (u'that', u'DT')), \\
((u'that', u'DT'), u'prep', (u'of', u'IN')), \\
((u'of', u'IN'), u'pobj', (u'course', u'NN')), \\
((u'loss', u'NN'), u'cop', (u'is', u'VBZ')), \\
((u'loss', u'NN'), u'nn', (u'language', u'NN'))\textbf{]}}

Then we can count the dependency words in one tuple that are both from the left half or the right half, e.g., \emph{`indicator'} depends on \emph{`the'} and both belong to the left half.

\paragraph{Japanese Parsing} For Japanese parsing we use J.DepP\footnote{\url{http://www.tkl.iis.u-tokyo.ac.jp/~ynaga/jdepp/}}:

\begin{CJK*}{UTF8}{gbsn}
CASE: \emph{``これ ら の 要素 と 予測 精度 の 特性 に つ い て 説明 し た 。"}

PARSING:

\emph{* 0 1D@0.908514
これら  名詞,代名詞,一般,*,*,*,これら,コレラ,コレラ     B@0.000000
の      助詞,連体化,*,*,*,*,の,ノ,ノ    I@0.000000\\
* 1 4D@0.000000
要素    名詞,一般,*,*,*,*,要素,ヨウソ,ヨーソ    B@0.999910
と      助詞,格助詞,一般,*,*,*,と,ト,ト I@0.000000\\
* 2 3D@0.993463
予測    名詞,サ変接続,*,*,*,*,予測,ヨソク,ヨソク        B@0.999645
精度    名詞,一般,*,*,*,*,精度,セイド,セイド    I@0.028107
の      助詞,連体化,*,*,*,*,の,ノ,ノ    I@0.000000\\
* 3 4D@0.000000
特性    名詞,一般,*,*,*,*,特性,トクセイ,トクセイ        B@0.999907
について        助詞,格助詞,連語,*,*,*,について,ニツイテ,ニツイテ       I@0.000000\\
* 4 -1D@0.000000
説明    名詞,サ変接続,*,*,*,*,説明,セツメイ,セツメイ    B@0.999984
し      動詞,自立,*,*,サ変・スル,連用形,する,シ,シ      I@0.014534
た      助動詞,*,*,*,特殊・タ,基本形,た,タ,タ   I@0.000878
。      記号,句点,*,*,*,*,。,。,。      I@0.001575}

The two ids at the begging of each line shows the dependency words. In this case, the last token is ``。" with id 4, we simply remove this token when counting the number.
\end{CJK*}

\bibliographystyle{acl_natbib_nourl}

\end{document}